\documentclass[10pt,twocolumn,letterpaper]{article}

\usepackage[pagenumbers]{cvpr} 

\usepackage{xcolor}         
\usepackage{adjustbox}
\usepackage{amsmath}
\usepackage{amsthm}
\usepackage{amsfonts,amssymb} 
\usepackage{graphicx}
\usepackage{algorithm}
\usepackage{algorithmic}
\usepackage{setspace}
\usepackage{multirow}
\usepackage{pifont}
\usepackage{bbding}
\usepackage{color}
\usepackage{wrapfig}
\usepackage{bm}
\usepackage{colortbl}

\definecolor{cvprblue}{rgb}{0.21,0.49,0.74}
\usepackage[pagebackref,breaklinks,colorlinks,allcolors=cvprblue]{hyperref}

\newcommand{\siyuan}[1]{{\color{black}#1}} 
\newcommand{\lanqing}[1]{{\color{black}#1}} 
\newcommand{\bihan}[1]{\textcolor{black}{#1}} 

\def\paperID{11059} 
\def\confName{CVPR}
\def\confYear{2025}

\newcommand{\N}     {\mathcal{N}}
\newcommand{\R}     {\mathcal{R}}

\newcommand{\Ndis}     {\mathcal{N}(\bm{0}, \mathbf{I} )}

\newcommand{\y}     {\bm{y}}

\newcommand{\ep}     {\bm{\epsilon}}
\newcommand{\x}     {\bm{x}}
\newcommand{\z}     {\bm{z}}
\newcommand{\n}     {\bm{n}}

\DeclareMathOperator*{\argmin}{argmin}

\def\ie{\emph{i.e.,}}
\def\eg{\emph{e.g.,}}

\title{Reconciling Stochastic and Deterministic Strategies for Zero-shot Image Restoration using Diffusion Model in Dual}

\author{Chong Wang$^1$, Lanqing Guo$^2$, Zixuan Fu$^1$, Siyuan Yang$^1$, Hao Cheng$^3$, Alex C. Kot$^1$, Bihan Wen$^1$\thanks{Bihan Wen is the corresponding author.}\\
$^1$Nanyang Technological University, Singapore\\ 
$^2$The University of Texas at Austin, USA~
$^3$Hebei University of Technology, China\\
}

\begin{document}
\maketitle

\begin{abstract}
%
%
%
%
\bihan{Plug-and-play (PnP) methods offer an iterative strategy for solving image restoration (IR) problems in a zero-shot manner, using a learned \textit{discriminative denoiser} as the implicit prior.}
More recently, a sampling-based variant of this approach, which utilizes a pre-trained \textit{generative diffusion model}, has gained great popularity for solving IR problems through stochastic sampling.
%
\bihan{The IR results using PnP with a pre-trained diffusion model demonstrate distinct advantages compared to those using discriminative denoisers, \ie improved perceptual quality while sacrificing the data fidelity. The unsatisfactory results are due to the lack of integration of these strategies in the IR tasks.}
\bihan{In this work, we propose a novel zero-shot IR scheme, dubbed Reconciling Diffusion Model in Dual (RDMD), which leverages only a \textbf{single} pre-trained diffusion model to construct \textbf{two} complementary regularizers.}
%
\bihan{Specifically, the diffusion model 
in RDMD will iteratively perform deterministic denoising and stochastic sampling, aiming to achieve high-fidelity image restoration with appealing perceptual quality.}
%
%
\bihan{RDMD also allows users to customize the distortion-perception tradeoff with a single hyperparameter, enhancing the adaptability of the restoration process in different practical scenarios. }
%
Extensive experiments on several IR tasks demonstrate that our proposed method could achieve superior results compared to existing approaches on both the FFHQ and ImageNet datasets.
Code is available at \url{https://github.com/ChongWang1024/RDMD}.
\end{abstract}

\begin{figure} [t]
  \centering 
    \includegraphics[width=0.99\linewidth]{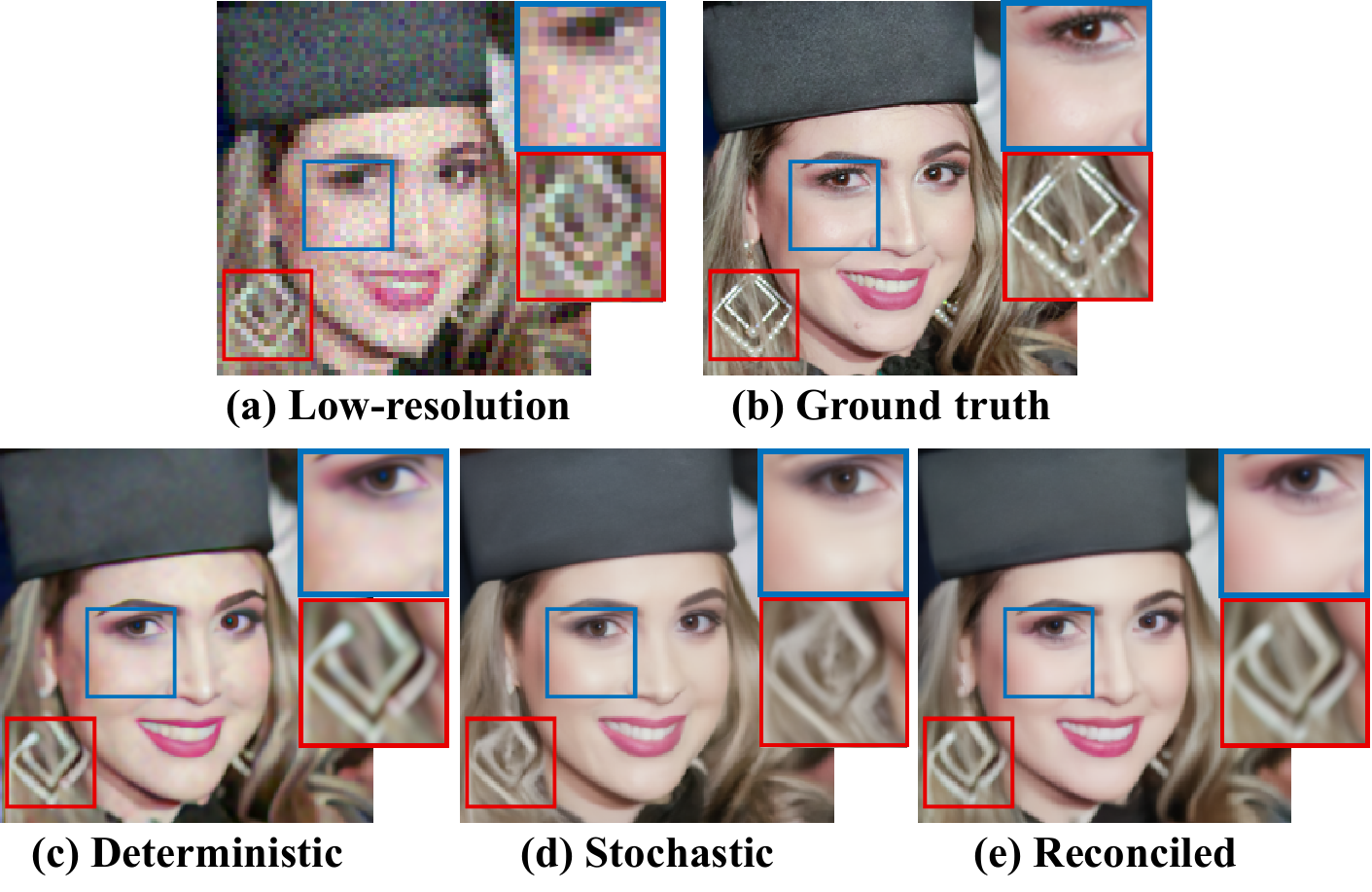}
\caption{An example of super-resolution (4$\times$) with noise level 0.05. 
From (a) to (e): (a) low-resolution input, (b) original image, (c) ours result via pure deterministic regression, (d) ours result via pure stochastic sampling, (e) ours by reconciling these two variants, respectively.
%
}
    \label{fig_intro}
\end{figure} 

\section{Introduction}
Image restoration (IR) aims to recover the latent clean image $\x$ from its degraded measurements $\y$.
This degradation process is typically modeled as $\y = A \x + \n$, where $A$ represents the forward model and $\n$ denotes the additive noise.
Different IR tasks correspond to specific observation models $A$. 
For example, in image inpainting, $A$ functions as a downsampling matrix, while in super-resolution tasks, $A$ acts as a decimation operator, which involves blurring followed by downsampling.
Directly restoring a clean image $\x$ from $\y$ becomes an ill-posed problem due to the underdetermined nature of the forward model $A$.
A common approach to address this restoration problem is to \siyuan{employ} 
a maximum a posteriori (MAP) optimization, formulated as:
\begin{equation}\small \label{eq_prob_map}
    \min_{\x}  \frac{1}{2\sigma_n^2} \lVert \y - A \x \rVert^2_2 + \lambda \R(\x),
\end{equation}
where $\R(\cdot)$ is the regularizer that encapsulates the image prior or distribution of the underlying clean images with the strength controlled by $\lambda$ and $\sigma_n$ is the noise level of additive Gaussian noise.
%
Recent deep learning-based methods have achieved impressive restoration results using the power of a large collection of data.
%
Although these task-specific networks excel in narrowly defined problems, they frequently struggle with generalizability across various IR tasks.
Consequently, there is a growing interest \siyuan{ in exploring zero-shot approaches to address these generalizability challenges.}

\bihan{One of the most popular approaches for zero-shot IR is the Plug-and-Play (PnP) scheme~\cite{venkatakrishnan2013plug}, which has demonstrated great flexibility in solving a wide range of IR problems without \siyuan{the need of} re-tuning for each specific task.}
%
With the aid of variable splitting algorithms,
the PnP approach is able to incorporate a pre-trained image denoiser as a replacement of the proximal mapping of the proximal operator corresponding to the regularizer in alternating iterations.
%
The PnP method can restore clean images by utilizing the implicit prior in a discriminatively learned denoiser, effectively removing the degradation through \textit{\textbf{deterministic}} regression.
More recently, with the rise of generative modeling, the denoising diffusion model (DDM)~\cite{ho2020denoising} has achieved great success in synthesizing images with high perceptual quality.
%
Therefore, leveraging such generative diffusion priors to conduct \textit{\textbf{stochastic}} sampling for IR has further surged the popularity of PnP-based strategies.
%

%
%
\bihan{In practice, the deterministic and stochastic approaches possess distinct strengths for solving image restoration problems. }
%
\bihan{Figure~\ref{fig_intro} shows one example of image super-resolution using our scheme: the result by applying deterministic regression (c) achieves high data fidelity while many unrealistic smearing artifacts are presented; Conversely, stochastic sampling (d) can enhance perceptual quality but often at the expense of fidelity to the original image.} 
%
%
\bihan{It demonstrates the complementariness of these two approaches in solving the ill-posed IR problems~\cite{wen2020set,zha2023multiple}, thus, the unsatisfactory results are due to the lack of integration of these strategies.}
Several attempts~\cite{lin2023diffbir, wang2023dr2, shang2024resdiff} have sought to combine these approaches to capitalize on their collective merits.
\bihan{Most of these methods simply cascade diffusion models with denoising networks for supervised learning, usually resulting in inconsistencies or mismatches in details after fusion, owing to the differing characteristics emphasized by each model. Moreover, employing a series of cascaded networks for IR tasks will lead to increased computational complexity and overfitting issues.}
%
%
%
%
%

An important question that arises here is: \textit{can we extend such an idea to zero-shot image restoration without inheriting its drawbacks?}
To circumvent these issues, we devise a unified iterative zero-shot framework, dubbed Reconciling Diffusion Model in Dual (RDMD), which integrates the merits of both stochastic and deterministic strategy, while only necessitating a single pre-trained diffusion model.
To the best of our knowledge, \siyuan{this is the first attempt to apply such a method to zero-shot IR.}
%
Specifically, the restoration process is performed via jointly deterministic regression and stochastic sampling along each iteration\siyuan{, without the need for additional networks.}
%
The proposed iterative strategy rests on the observation that the training process of the diffusion model can be interpreted as discriminatively training a non-blind Gaussian denoiser.
Thus, we can leverage a single pre-trained diffusion model while constructing two complementary regularizers.
%
We further derive a rigorous formulation of the proposed framework, where the contribution of these two regularizers can be controlled easily through a weighting parameter.
Building on this \siyuan{framework}, our method can provide flexible customization on the distortion-perception tradeoff\siyuan{, enhancing its suitability for various distinct IR tasks.} 
%
%
Experimental results show that the proposed method consistently achieves superior performance in various restoration tasks on the FFHQ and ImageNet datasets.

The main contributions of this work are as follows:
\begin{itemize}
    \item We first demonstrate that a single pre-trained diffusion model can be used to construct two complementary regularizers\siyuan{, \ie stochastic and deterministic strategies.}
    \item Building on this, we conduct the first attempt to integrate the deterministic and stochastic strategies in a unified framework that solves zero-shot IR problems iteratively.
    \item Comprehensive experimental results on several IR tasks demonstrate our methods outperform existing competing approaches on both the FFHQ and ImageNet datasets.
\end{itemize}

\section{Background and Related Work}
\subsection{Plug-and-play for Image Restoration}
PnP methods~\cite{venkatakrishnan2013plug} have attracted significant attention due to their unique flexibility for solving different image restoration problems without the need for re-training on each task.
The first study of PnP can be traced back to~\cite{venkatakrishnan2013plug} where the authors utilized various denoisers, such as K-SVD~\cite{aharon2006k} or BM3D~\cite{dabov2007image} as a replacement for the proximal operator corresponding to the regularizer in ADMM~\cite{boyd2011distributed} iterations.
Apart from ADMM, other variable splitting algorithms, such as half quadratic splitting~\cite{hqs}, have been adopted in the PnP paradigms~\cite{zhang2017learning, zhang2021plug, wang2024progressive, wang2023deep}.
Specifically, The optimization in~\eqref{eq_prob_map} can be solved by introducing an auxiliary variable $\z$ as:
\begin{equation}\small
        \min_{\z, \x} \frac{1}{2\sigma_n^2}  \lVert  \y - A \z \rVert^2_2 +  \frac{\mu}{2}  \lVert \z - \x \rVert^2_2 + \lambda \R(\x),
\end{equation}
where $\mu$ is the penalty parameter of the Lagrangian term.
Then it can be decoupled and optimized via alternation between two steps:
\small
\begin{align}\label{eq_hqs_x}
    &\x_t = \argmin_{\x}  \frac{\mu}{2}  \lVert \x - \z_{t} \rVert^2_2 + \lambda \R(\x),\\ \label{eq_hqs_z}
    &\z_{t-1} = \argmin_{\z}   \lVert  \y - A \z \rVert^2_2 +  {\mu}  \lVert \z - \x_{t} \rVert^2_2.
\end{align}
\normalsize
With the merits of such a splitting algorithm, the data term $\lVert \y - A \x \rVert^2_2$ and prior term $\R(\x)$ can be independently addressed as two separate sub-problems.
%
The data consistency step in~\eqref{eq_hqs_z} prevents the restored images from violating the information in measurements $\y$ and forward model $A$.
Besides, the optimization in~\eqref{eq_hqs_z} is a least-squares problem with a quadratic penalty term, which has a closed-form solution.
The update process for $\x_t$ corresponds to a Gaussian denoising problem that can be solved using any denoiser.
%
With the emergence of deep learning-based approaches in recent decades, adopting a pre-trained deep denoiser as the solver for~\eqref{eq_hqs_x} has become a popular choice in the PnP framework.
%
For example, DPIR~\cite{zhang2021plug} adopts a non-blind CNN denoiser learned discriminative on a wide range of noise levels achieved promising restoration results when incorporated in the PnP framework.
A similar idea is also introduced in regularization-by-denoising (RED)~\cite{romano2017little} where the authors proposed an explicit formulation on the regularizer using a pre-trained Gaussian denoiser for image restoration.
One significant strength of RED is that it demonstrates a better convergence behavior compared to the PnP approaches.

\subsection{Denoising Diffusion Models for Image Restoration}
Denoising diffusion probabilistic models~\cite{ho2020denoising, sohl2015deep} have emerged as a powerful class of unconditional generative models, which has demonstrated impressive performance on high-quality image synthesis~\cite{nichol2021improved, ho2022cascaded, rombach2022high, dhariwal2021diffusion}. 
The success of diffusion models has motivated another line of research~\cite{zhu2023denoising, chung2023diffusion, chung2022improving, wang2022zero, song2024solving} by leveraging a pre-trained diffusion network as the generative prior to address image restoration problems in a zero-shot manner.
We follow the most popular formulation of DDM in~\cite{ho2020denoising}, where the diffusion model consists of two main processes.
In the forward process of DDM, a clean image $\x_0 \sim p(\x)$ from a desired distribution is gradually perturbed by a small Gaussian noise $\ep \sim \Ndis$:
\begin{equation}\small\label{eq_diff_forward}
    \x_t = \sqrt{1-\beta_t} \x_{t-1} + \sqrt{\beta_t} \ep,
\end{equation}
%
where $\{\beta_t\}_{t=1}^T$ is the noise schedule  in ascending order $0<\beta_1 < \dots < \beta_T < 1$.
Through reparameterization, a noisy data $\x_t$ could be sampled directly from $\x_0$ in a closed form:
%
$\x_t = \sqrt{\bar{\alpha}_t} \x_0 + \sqrt{1-\bar{\alpha}_t} \ep$
where $\alpha_t = 1 - \beta_t$ and $\bar{\alpha}_t = \prod_{i=1}^t \alpha_i$.
The DDM is trained to approximate the reverse process of~\eqref{eq_diff_forward}, progressing from a pure Gaussian noise $\x_T \sim \Ndis$ to a clean natural image $\x_0$ as:
\begin{equation}\small
    \x_{t-1} = \frac{1}{\sqrt{\alpha_t}} \left( \x_t - \frac{1-\alpha_t}{\sqrt{1 - \bar{\alpha}_t}} \ep_\theta (\x_t ; t)\right) + \beta_t \ep_t.
\end{equation}
The reverse process of DDM in~\cite{ho2020denoising} necessitates thousands of iterations, leading to a slow generation process for each image.
As a remedy,~\cite{song2020denoising} proposed denoising diffusion implicit models (DDIM) which define a non-Markovian process for generation as:
\begin{equation}\small
    \x_{t-1} = \sqrt{\bar{\alpha}_{t-1}} \x_{0|t} +  \hat{\sigma}_t \ep_\theta (\x_t; t) + \tilde{\sigma}_t \ep_t,
\end{equation}
where $\hat{\sigma}_t = \sqrt{1 - \bar{\alpha}_{t-1} - \tilde{\sigma}_t^2}$.
The parameter $\tilde{\sigma}_t$ controls the weight between predicted noise $\ep_\theta$ and pure Gaussian noise $\ep_t$, and $\x_{0|t}$ the intermediate prediction of clean image form $\x_t$ which takes the form:
\begin{equation}\small\label{eq_reverse_x0|t}
    \x_{0|t} = f_\theta (\x_t; t)= \frac{1}{\sqrt{\bar{\alpha}_t}} \left(\x_t - \sqrt{1-\bar{\alpha}_t} \ep_\theta (\x_t; t) \right),
\end{equation}
which can be viewed as a denoising process, denoted as $f_\theta$.
Existing approaches typically insert constraints on $\x_{0|t}$ to guide the generative trajectory to align with the information of the degradation model in each IR task.
For example, \cite{wang2022zero} performed a range-null space decomposition on $\x_{0|t}$ where the diffusion process is only performed on the null-space information.
A relaxed constraint is proposed in~\cite{zhu2023denoising}, where the authors incorporated a least-squares penalty of the data term $\lVert \y - A \x \rVert^2_2$ to guarantee the data consistency.
Another study~\cite{song2024solving} adopted a latent diffusion model while ensuring data consistency by optimizing $\x_{0|t}$ that minimizes the data term during the reverse sampling process.

\begin{figure*} [t]
  \centering 
    \includegraphics[width=0.98\linewidth]{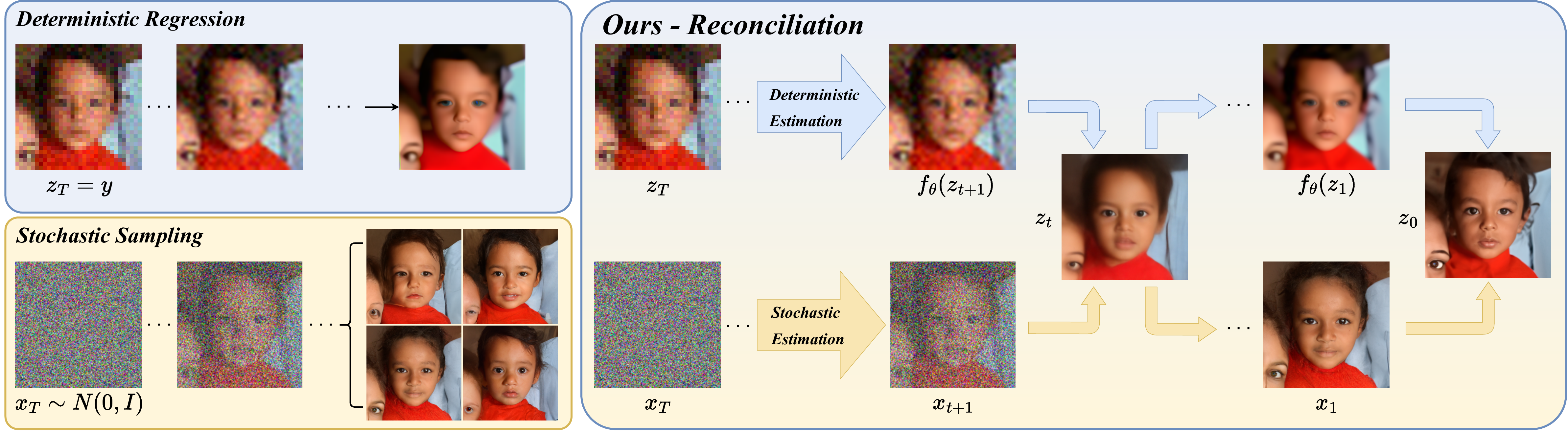}
\caption{Comparison of restoration processes: deterministic regression, stochastic sampling, and our proposed reconciliation approach. 
The deterministic regression method (top left) generates high-fidelity restorations through pointwise estimations. The stochastic sampling method (bottom left) produces diverse outputs by sampling from a learned distribution, enhancing perceptual quality. 
Our reconciliation approach (right) unifies deterministic estimation for accurate details with stochastic estimation for diverse possibilities, achieving both high fidelity and perceptual richness in the final restored images.}
    \label{fig_dual_demo}
  \end{figure*}

\section{Method}
Classical plug-and-play methods incorporate \siyuan{either} a predefined denoiser~\cite{dabov2007image, chan2016plug} or a pre-trained deep denoiser~\cite{zhang2017learning, wang2022repnp} to approximate the solution to~\eqref{eq_hqs_x}.
This approach enables a deterministic reconstruction for each image restoration problem through iterative regression with an off-the-shelf denoiser.
%
%
An alternative line of research tackles the optimization in~\eqref{eq_prob_map} using a generative diffusion prior from the perspective of stochastic sampling~\cite{zhu2023denoising, wang2022zero}, leading to results with appealing perceptual quality in a zero-shot manner.
Different from previous approaches, we propose a unified framework that \siyuan{integrates} the merits of \siyuan{both deterministic regression and stochastic sampling for restoration.}
A comparison of these frameworks is illustrated in Figure~\ref{fig_dual_demo}.
%

\subsection{Overall Framework}\label{sec_framework}
Following the existing HQS algorithm, we introduce an auxiliary variable $\z$ while also performing partial replacement on the regularizer term, as follows:
\begin{equation}\small
        \min_{\z, \x}  \frac{1}{2\sigma_n^2} \lVert  \y - A \z \rVert^2_2 + \frac{\mu}{2}   \lVert \z - \x \rVert^2_2 + \tau\lambda \R_\theta(\x) + (1-\tau)\lambda \R_\theta(\z),
\end{equation}
where $\R_\theta(\cdot)$ is the regularization term parameterized by a trained deep network with parameters $\theta$.
Then it can be split into two sub-problems:
\small
\begin{align}\label{eq_our_x}
    &\x_t = \argmin_{\x}  \frac{1}{2  (\sqrt{\tau \lambda / \mu})^2 }  \lVert \x - \z_{t} \rVert^2_2 +  \R^S_\theta(\x),\\ 
    \begin{split}\label{eq_our_z}
    &\z_{t-1} = \argmin_{\z} \frac{1}{2}  \lVert  \y - A \z \rVert^2_2 + \frac{\mu \sigma_n^2}{2}  \lVert \z - \x_{t} \rVert^2_2 \\
    & \qquad \qquad \qquad \qquad \qquad  + (1-\tau)\lambda \sigma_n^2 \R^D_\theta(\z).
    \end{split}
\end{align}
\normalsize
One may note that the main difference of our proposed framework is that the regularizer $\R_\theta(\cdot)$ is involved in both iterative processes, 
\lanqing{with its strength controlled by a weighting coefficient $\tau$.}
However, directly incorporating an identical regularizer in both sub-problems may lead to a trivial solution.
%
Building upon this formulation, we construct two distinct regularization terms $\R^{S}_\theta(\cdot)$ and $\R^D_\theta(\cdot)$, both derived from a single pre-trained diffusion model $\ep_\theta$.
These regularization terms are tailored to emphasize different characteristics essential for the respective updates of $\x$ and $\z$.
Specifically, this framework enables simultaneous stochastic sampling and deterministic regression throughout the iterations, leveraging the dual strategies on one diffusion model.

\begin{algorithm}[t]
\caption{RDMD} 
\label{alg1}
\begin{algorithmic}
\REQUIRE $\y$, $\N_{\theta}$, $\lambda$, $\eta$, $\tau$, $\zeta$,$\{\sigma'_t\}^T_{t=1}$ defined in Section~\ref{sec_framework}.\\
\textbf{Initialization:}  $\x_T \sim \Ndis$, $\z_T=A^\top \y$
\FOR{$t = T,\dots ,1$}
\STATE \textit{\textcolor{gray}{\# pre-calculate $t'$ for deterministic resgression}}
\STATE $t' = \argmin_{i\in [1, T]} |\sigma'_t - \frac{\sqrt{1-\bar{\alpha}_i}}{\sqrt{\bar{\alpha}_i}}| $
\STATE \textit{\textcolor{gray}{\# stochastic estimation}}
\STATE $\x_{0|t} = \frac{1}{\sqrt{\bar{\alpha}_t}} \left(\x_t - \sqrt{1-\bar{\alpha}_t} \ep_\theta (\x_t; t) \right)$
\STATE \textit{\textcolor{gray}{\# deterministic estimation}}
\STATE $f_\theta(\z_t; {t'}) = \frac{1}{\sqrt{\bar{\alpha}_{t'}}} \left(\z_t - \sqrt{1-\bar{\alpha}_{t'}} \ep_\theta (\z_t; {t'}) \right)$
\STATE \textit{\textcolor{gray}{\# main update process}}
\STATE $\z_{t-1} = \z_{t} - \eta \big( \nabla_{\z_t} \lVert  \y - A \z_t \rVert^2_2 + \tau \lambda\frac{\sigma_n^2}{\bar{\sigma}_t^2} (\z_{t} - \x_{0|t})$
\STATE $\qquad \qquad \qquad \qquad \quad + (1-\tau) \lambda \sigma_n^2 (\z_t - f_\theta(\z_t;t'))  \big)$
\STATE \textit{\textcolor{gray}{\# renoising}}
\STATE $\hat{\ep}_t = \frac{1}{\sqrt{1-\bar{\alpha}_t}}(\x_t - \sqrt{\bar{\alpha}_t}\z_{t-1})$ 
\STATE $\ep_t \sim \N (\bm{0}, \bm{I})$
\STATE $\x_{t-1} = \sqrt{\bar{\alpha}_{t-1}} \z_{t-1} + \sqrt{1-\bar{\alpha}_{t-1}}(\sqrt{1-\zeta}\hat{\ep}_t+\sqrt{\zeta}\ep_t)$ 
\ENDFOR
\STATE $\textbf{return} \;\x_0$
\end{algorithmic}
\end{algorithm}

\noindent\textbf{Stochastic sampling in $\x$ sub-problem.}
%
Existing PnP approaches typically replace the proximal mapping in~\eqref{eq_our_x} with a discriminatively trained denoiser in each iteration.
Instead, we approximate the solution using a single step of reverse diffusion in~\eqref{eq_reverse_x0|t} to fully exploit the generative \siyuan{capabilities} 
of the diffusion model.
From the Bayesian perspective,~\eqref{eq_our_x} corresponds to a Gaussian denoising problem with noise sigma $\sqrt{\tau \lambda / \mu}$.
%
\siyuan{By setting $\sqrt{\tau \lambda / \mu}=\bar{\sigma}_t = \sqrt{\frac{1 - \bar{\alpha}_t}{\bar{\alpha}_t}}$, we establish a connection between~\eqref{eq_our_x} and~\eqref{eq_reverse_x0|t}.}
\siyuan{This alignment ensures that the denoising strength of this step corresponds with the noise variance schedule of the pre-trained diffusion model.}
%
Moreover, the estimation $\x_{0|t}$ in the reverse diffusion process~\eqref{eq_reverse_x0|t} is closely related to the classical results of Tweedie's formula~\cite{robbins1992empirical, stein1981estimation, efron2011tweedie} as $\x_{0|t} \simeq \x_t + \bar{\sigma}_t^2 \ep_\theta(\x_t; t)$, which is equivalent to a denoising process with noise level $\bar{\sigma}_t$.
\siyuan{Therefore, this approach can be seamlessly integrated into the iterative process in~\eqref{eq_our_x} as follows:}
%
\small
\begin{align}
    &\x_{0|t} = \frac{1}{\sqrt{\bar{\alpha}_t}} \left(\x_t - \sqrt{1-\bar{\alpha}_t} \ep_\theta (\x_t; t) \right),\\ 
    \begin{split}
    &\z_{t-1} = \argmin_{\z} \frac{1}{2}  \lVert  \y - A \z \rVert^2_2 + \frac{\mu \sigma_n^2}{2}  \lVert \z - \x_{0|t} \rVert^2_2 \\
    & \qquad \qquad \qquad \qquad \qquad \quad + (1-\tau)\lambda \sigma_n^2 \R^D_\theta(\z),
    \end{split}\\ \label{eq_diffpir_sampling}
    &\x_{t-1} = \sqrt{\bar{\alpha}_{t-1}} \z_{t-1} + \sqrt{1-\bar{\alpha}_{t-1}}(\sqrt{1-\zeta}\hat{\ep}_t+\sqrt{\zeta}\ep_t).
\end{align}
\normalsize
%
\siyuan{The update process in~\eqref{eq_diffpir_sampling} reintroduces noise to the current estimate $\z_{t-1}$ to maintain alignment with the diffusion process.}
Following~\cite{zhu2023denoising}, 
\siyuan{rather than} directly adding the predicted noise $\ep_\theta(\x_t;t)$, we use the effective noise based on the relationship in~\eqref{eq_reverse_x0|t} as:
$\hat{\ep}_t = \frac{1}{\sqrt{1-\bar{\alpha}_t}} (\x_t - \sqrt{\bar{\alpha}_t} \z_{t-1}).$
%
Besides, $\zeta$ is the weighting parameter to control the stochasticity by balancing the effective noise $\hat{\ep}_t$ and the pure Gaussian noise $\ep_t$.
Based on this, we \siyuan{construct} 
the update of $\x_t$ to perform diffusion-based stochastic sampling.

\begin{figure} [t]
  \centering 
    \includegraphics[width=0.9\linewidth]{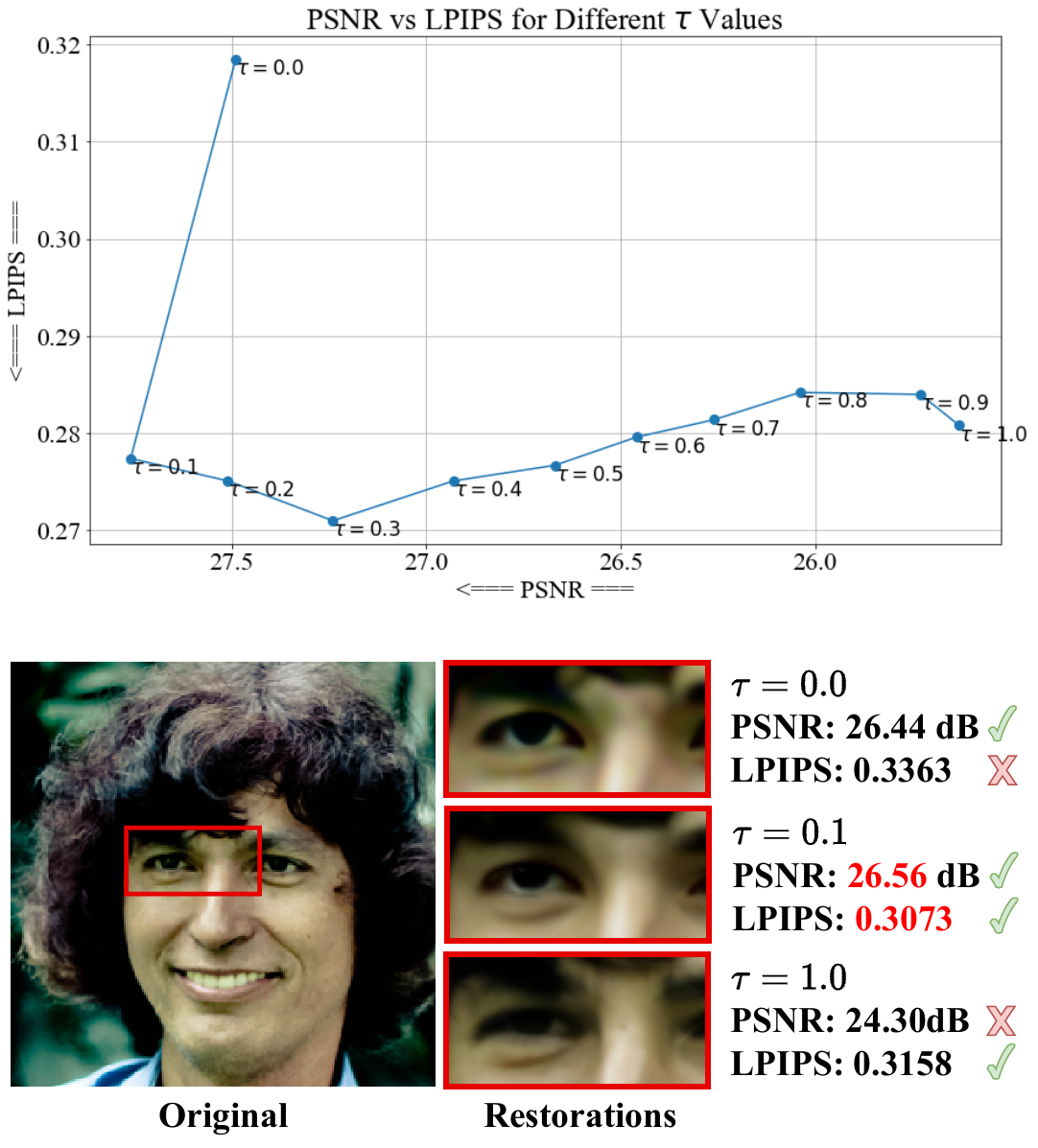}
\caption{An example of controlling distortion-perception tradeoff for 4$\times$ super-resolution via adjusting different values of $\tau$.
Specifically, $\tau \to 0$ leads to deterministic regression while $\tau \to 1$ promotes the stochasticity.}
    \label{fig_tradeoff}
  \end{figure}

\noindent\textbf{Deterministic regression in $\z$ sub-problem.} 
The training process of the diffusion model in~\cite{ho2020denoising} is to predict the noise used to perturb the clean image $\x_0$\siyuan{, formulated} as:
\begin{equation}\small
    \min_\theta \mathbb{E}_{\ep \sim \Ndis}  \lVert \ep - \ep_\theta (\sqrt{\bar{\alpha}_t} \x_0 + \sqrt{ 1- \bar{\alpha}_t}\ep ; t)\rVert^2_2,
\end{equation}
which can be interpreted as training a non-blind Gaussian denoiser where the noise level is implicitly embedded via time step $t$.
Therefore, such a pre-trained diffusion model \siyuan{can seamlessly integrate} 
into the PnP framework as the regularizer for deterministic regression.
However, by introducing the regularizer in the update process of $\z_t$, there is no closed-form solution to~\eqref{eq_our_z}.
Alternatively, we can still approximate the solution via gradient methods as:
\begin{equation}\small
\begin{split}
    \z_{t-1} = \z_{t} - \eta \big(  \nabla_{\z_t} \lVert  \y &- A \z_t \rVert^2_2 + \mu \sigma_n^2 (\z_{t} - \x_{0|t}) \\
    & \quad + (1-\tau) \lambda \sigma_n^2 \nabla \R_\theta^D(\z_{t})  \big),
\end{split}
\end{equation}
where the explicit formulation of the regularizer gradient is required.
Existing PnP methods typically employ a denoiser to replace the proximal mapping associated with a regularizer,
\siyuan{but lack the capability to directly model its derivative.}
%
In contrast, Regularization by denoising~\cite{romano2017little} allows for an explicit regularization functional parameterized using a denoiser network.
Thus, we construct our deterministic regularizer in~\eqref{eq_our_z} using the pre-trained diffusion model as:
\begin{equation}\small
\R_\theta^D(\z_t) = \frac{1}{2}\z_t^\top (\z_t - f_\theta(\z_t; t')),
\end{equation}
where $f_\theta$ is the denoising process formed by $\ep_\theta$ as in~\eqref{eq_reverse_x0|t} and $t'$ is the equivalent time step according to the pre-defined noise level schedule $\{\sigma'_t\}^T_{t=1}$ on $\z_t$.
Under several assumptions, \ie symmetric Jacobian~\cite{romano2017little, reehorst2018regularization}, the gradient of RED has an explicit expression as $\nabla \R(\z_t) = \z_t - f_\theta(\z_t; t')$.
Therefore, following this formulation, we could obtain a deterministic regression process~\eqref{eq_our_z} without introducing extra deep models for regularization.
The overall iterative framework can be illustrated in Algorithm~\ref{alg1}, dubbed as \textit{\textbf{Reconciling Diffusion Model in Dual (RDMD)}}.

\begin{table*}
    \centering
    \caption{Quantitative evaluation results (PSNR, SSIM, and LPIPS) of image restoration tasks: \textit{Gaussian} deblur, \textit{motion} deblur, and super-resolution with $\sigma_n$ = 0.05 on the FFHQ 256$\times$256 dataset. \textbf{Bold}: best, \underline{underline}: second best.}
    \setlength{\tabcolsep}{0.5mm}
    \adjustbox{width=\linewidth}{
    \begin{tabular}{lccccccccccccccc}
    \toprule
         \textbf{FFHQ}&\multicolumn{3}{c}{\textbf{Deblur (\textit{Gaussian})}}&\multicolumn{3}{c}{\textbf{Deblur (\textit{motion})}}&\multicolumn{3}{c}{\textbf{Super-resolution (2$\times$)}}& \multicolumn{3}{c}{\textbf{Super-resolution (4$\times$)}}& \multicolumn{3}{c}{\textbf{Super-resolution (8$\times$)}}\\
     \midrule
        Method &\small{PSNR$\uparrow$}&\small{SSIM$\uparrow$}&\small{LPIPS$\downarrow$}&\small{PSNR$\uparrow$}&\small{SSIM$\uparrow$}&\small{LPIPS$\downarrow$}&\small{PSNR$\uparrow$}&\small{SSIM$\uparrow$}&\small{LPIPS$\downarrow$}&\small{PSNR$\uparrow$}&\small{SSIM$\uparrow$}&\small{LPIPS$\downarrow$}&\small{PSNR$\uparrow$}&\small{SSIM$\uparrow$}&\small{LPIPS$\downarrow$}\\
        \midrule
         MCG\cite{chung2022improving}&11.19&0.0968&0.8042&11.33&0.0831&0.8088&19.57&0.3482&0.5648&17.34&0.2153&0.7190&16.81&0.1829&0.7676\\
         DPS\cite{chung2023diffusion}&23.78&0.6776&0.2829&20.64&0.5732&0.3461&25.06&0.7418&0.2738&22.85&0.6503&0.3215&20.54&0.5597&0.3726\\
         Resample\cite{song2024solving}&23.50&0.4962&0.4612&18.60&0.2566&0.6455&23.07&0.4754&0.4828&18.28&0.2563&0.6691&18.74&0.2712&0.6807\\
         BIRD\cite{chihaoui2024blind}&26.14&0.7009&0.2928&24.21&0.6419&0.3412&27.04&0.6150&0.3310&22.51&0.3870&0.5190&19.67&0.3070&0.6340\\
         \midrule
          RED\cite{romano2017little}\footnotesize{(Ours - deterministic)}&\underline{27.84}&\textbf{0.8096}&0.3170&26.09&0.7372&0.3613&27.87&0.7919&0.2562&\underline{27.05}&\underline{0.7960}&0.3277&\underline{22.22}&\underline{0.6431}&0.4883\\
           DiffPIR\cite{zhu2023denoising}\footnotesize{(Ours - stochastic)}&27.30&0.7672&\textbf{0.2372}&\underline{26.63}&\underline{0.7399}&\textbf{0.2529}&\underline{29.39}&\underline{0.8060}&\underline{0.2172}&26.27&0.7332&\textbf{0.2674}&21.08&0.5634&\underline{0.3666}\\
         \midrule
         \cellcolor{gray!20}Ours&\cellcolor{gray!20}\textbf{28.29}&\cellcolor{gray!20}\underline{0.8057}&\cellcolor{gray!20}\underline{0.2770}&\cellcolor{gray!20}\textbf{27.05}&\cellcolor{gray!20}\textbf{0.7776}&\cellcolor{gray!20}\underline{0.2767}&\cellcolor{gray!20}\textbf{31.03}&\cellcolor{gray!20}\textbf{0.8715}&\cellcolor{gray!20}\textbf{0.1807}&\cellcolor{gray!20}\textbf{27.84}&\cellcolor{gray!20}\textbf{0.8012}&\cellcolor{gray!20}\underline{0.2768}&\cellcolor{gray!20}\textbf{23.31}&\cellcolor{gray!20}\textbf{0.6524}&\cellcolor{gray!20}\textbf{0.3614}\\ 
      \bottomrule
    \end{tabular}
    }
    \label{tab_ffhq}
\end{table*}

\begin{table*}
    \centering
    \caption{Quantitative evaluation results (PSNR, SSIM, and LPIPS) of image restoration tasks: \textit{Gaussian} deblur, \textit{motion} deblur, and super-resolution with $\sigma_n$ = 0.05 on the ImageNet 256$\times$256 dataset. \textbf{Bold}: best, \underline{underline}: second best.}
    \setlength{\tabcolsep}{0.5mm}
    \adjustbox{width=\linewidth}{
    \begin{tabular}{lccccccccccccccc}
    \toprule
         \textbf{ImageNet}&\multicolumn{3}{c}{\textbf{Deblur (\textit{Gaussian})}}&\multicolumn{3}{c}{\textbf{Deblur (\textit{motion})}}&\multicolumn{3}{c}{\textbf{Super-resolution (2$\times$)}}& \multicolumn{3}{c}{\textbf{Super-resolution (4$\times$)}}& \multicolumn{3}{c}{\textbf{Super-resolution (8$\times$)}}\\
     \midrule
        Method &\small{PSNR$\uparrow$}&\small{SSIM$\uparrow$}&\small{LPIPS$\downarrow$}&\small{PSNR$\uparrow$}&\small{SSIM$\uparrow$}&\small{LPIPS$\downarrow$}&\small{PSNR$\uparrow$}&\small{SSIM$\uparrow$}&\small{LPIPS$\downarrow$}&\small{PSNR$\uparrow$}&\small{SSIM$\uparrow$}&\small{LPIPS$\downarrow$}&\small{PSNR$\uparrow$}&\small{SSIM$\uparrow$}&\small{LPIPS$\downarrow$}\\
        \midrule
         MCG\cite{chung2022improving}&11.11&0.0982&0.7338&11.05&0.0362&0.7535&15.12&0.2217&0.6259&16.31&0.2150&0.6727&15.87&0.1507&0.7271\\
         DPS\cite{chung2023diffusion}&20.09&0.4774&0.4272&20.14&0.4871&0.4336&21.31&0.5514&0.4392&19.47&0.4510&0.4961&17.65&0.3522&\underline{0.5505}\\
         Resample\cite{song2024solving}&18.88&0.2957&0.5579&15.49&0.1857&0.6571&21.01&0.4674&0.4334&16.46&0.2235&0.6163&17.55&0.2207&0.6968\\
         BIRD\cite{chihaoui2024blind}&22.04&0.5360&0.4107&20.79&0.4976&0.4499&24.94&0.6841&0.3047&21.37&0.4521&0.5001&19.67&0.3073&0.6334\\
         \midrule
         RED\cite{romano2017little}\footnotesize{(Ours - deterministic)}&\underline{23.24}&\textbf{0.6090}&0.4282&22.52&0.5674&0.4963&\underline{25.87}&\underline{0.7199}&0.3023&\underline{23.16}&\underline{0.6027}&0.4324&\underline{20.42}&\underline{0.4634}&0.6024\\
          DiffPIR\cite{zhu2023denoising}\footnotesize{(Ours - stochastic)}&21.97&0.5363&\textbf{0.3827}&\textbf{23.34}&\textbf{0.6276}&\textbf{0.3736}&25.33&0.7092&\underline{0.2900}&22.80&0.5967&\textbf{0.3871}&19.44&0.4292&\textbf{0.5411}\\
         \midrule
         \cellcolor{gray!20}Ours&\cellcolor{gray!20}\textbf{23.30}&\cellcolor{gray!20}\underline{0.6004}&\cellcolor{gray!20}\underline{0.4079}&\cellcolor{gray!20}\underline{22.77}&\cellcolor{gray!20}\underline{0.6037}&\cellcolor{gray!20}\underline{0.4247}&\cellcolor{gray!20}\textbf{25.94}&\cellcolor{gray!20}\textbf{0.7563}&\cellcolor{gray!20}\textbf{0.2648}&\cellcolor{gray!20}\textbf{23.28}&\cellcolor{gray!20}\textbf{0.6203}&\cellcolor{gray!20}\underline{0.4000}&\cellcolor{gray!20}\textbf{20.83}&\cellcolor{gray!20}\textbf{0.4736}&\cellcolor{gray!20}0.5690\\ 
      \bottomrule
    \end{tabular}
    }
    \label{tab_imagenet}
\end{table*}

\subsection{Discussion}\label{sec_discussion}
\noindent\textbf{Relevance to existing methods.}
Our algorithm represents a generalized framework that can extend several existing methods. 
We employ a single pre-trained diffusion model and construct two regularizers, emphasizing generative and discriminative, respectively.
The contribution of \siyuan{each regularizer} 
is controlled via the weighting parameter $\tau$.
%
\siyuan{At one extreme} $\tau=0$, the framework operates in a completely deterministic manner, utilizing the pre-trained diffusion model as a non-blind Gaussian denoiser within the RED iteration \siyuan{following} 
the steepest descent.
Conversely, when $\tau=1$, the framework shifts to a fully stochastic mode. In this scenario, our iterative process resembles DiffPIR, with the key distinction being our use of gradient descent for the data consistency step, whereas DiffPIR employs an analytic solution.
Thus, \siyuan{our unified framework effectively leverages the benefits of both deterministic regression and stochastic sampling.}

\noindent\textbf{Customizing the distortion-perception tradeoff.}
Diffusion models are renowned for their exceptional ability to synthesize images with outstanding perceptual quality.
However, these models \siyuan{often introduce} 
artificial details not present in the original images when striving for realism.
In image restoration, while perceptual quality is an important consideration, the focus often \siyuan{shift} 
to distortion metrics that \siyuan{measure} 
fidelity to the original image.
\siyuan{While deep models trained to minimize pixel-wise differences with the ground truth can achieve high fidelity, they tend to result in overly smooth patterns in the restored images.}
%
Existing studies~\cite{blau2018perception, liu2019classification, freirich2021theory} have proved that there is a fundamental conflict between distortion and perceptual quality.
Therefore, achieving a balance between perception and distortion in restoration results is crucial.
As mentioned above, we propose a unified framework that integrates deterministic regression and stochastic sampling, controlled by a single parameter $\tau$.
By adjusting $\tau$ within the range of $[0, 1]$, one can tailor the trade-off between perception and distortion to suit various \siyuan{restoration} challenges, enhancing the adaptability of the restoration process.
%
Figure~\ref{fig_tradeoff} \siyuan{illustrates how adjusting 
$\tau$ affects the tradeoff between fidelity and perceptual quality in 4$\times$ super-resolution.}
%
Lower $\tau$ values favor deterministic restoration, yielding higher fidelity, as reflected by higher PSNR values.
%
%
However, this can result in unrealistic artifacts, as seen when $\tau=0$, where smearing artifacts are evident despite the high PSNR.
On the contrary, when the algorithm runs in a pure stochastic manner ($\tau=1$), it leads to enhanced perceptual quality.
However, such high perceptual quality usually comes with a sacrifice of fidelity to the original image, as illustrated by the blackened left eye in Figure~\ref{fig_tradeoff}.
By reconciling these two strategies, such as with $\tau=0.1$, our method achieves a superior balance, optimizing both perceptual quality and fidelity.

\begin{figure*} [t]
  \centering 
    \includegraphics[width=\linewidth]{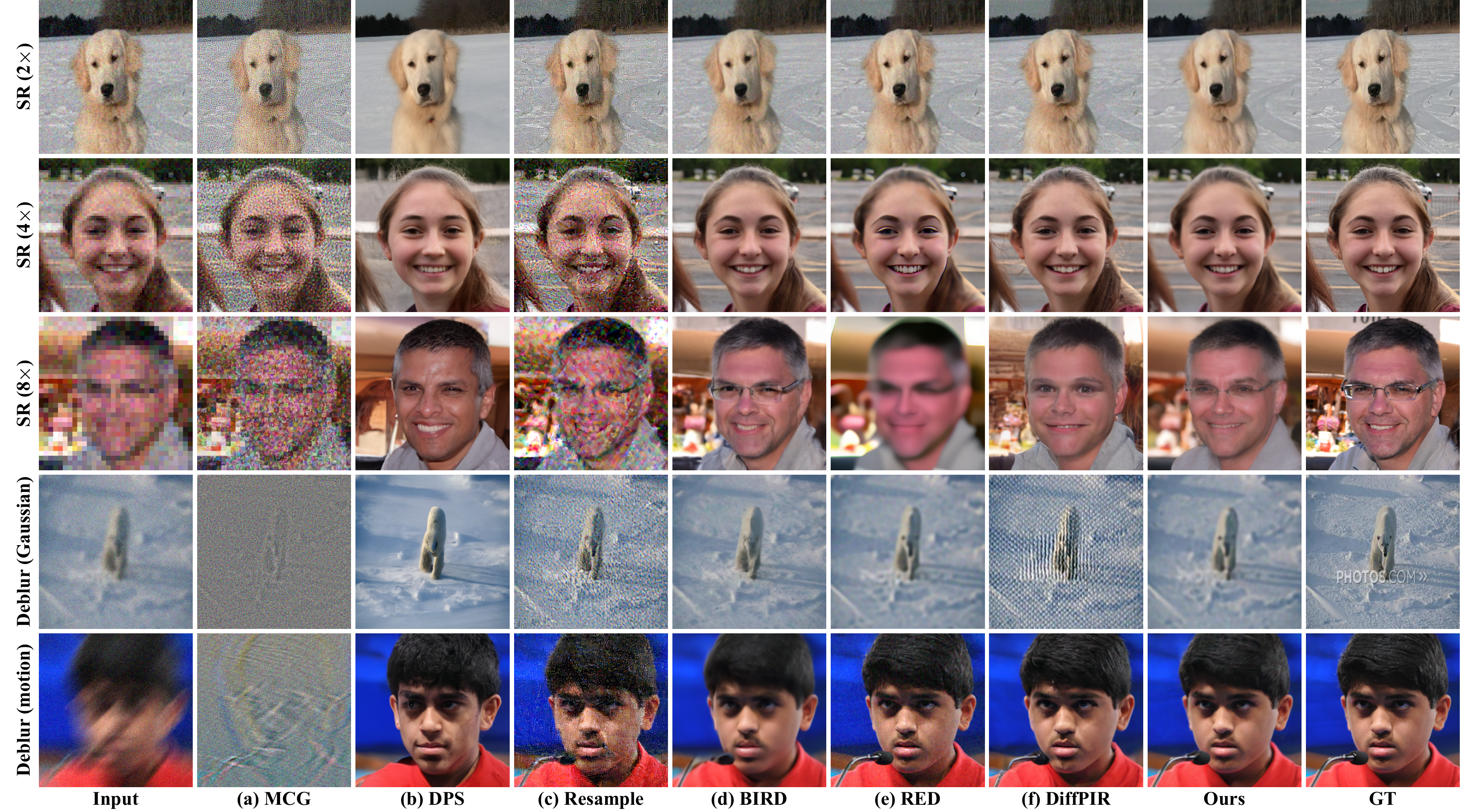}
\caption{Example of IR results of (a) MCG~\cite{chung2022improving}, (b) DPS~\cite{chung2023diffusion}, (c) Resample~\cite{song2024solving}, (d) BIRD~\cite{chihaoui2024blind}, (e) RED~\cite{romano2017little}, (f) DiffPIR~\cite{zhu2023denoising}, and Ours on FFHQ and  ImageNet datasets.}
    \label{fig_visual}
  \end{figure*} 

\begin{figure} [h]
  \centering 
    \includegraphics[width=0.97\linewidth]{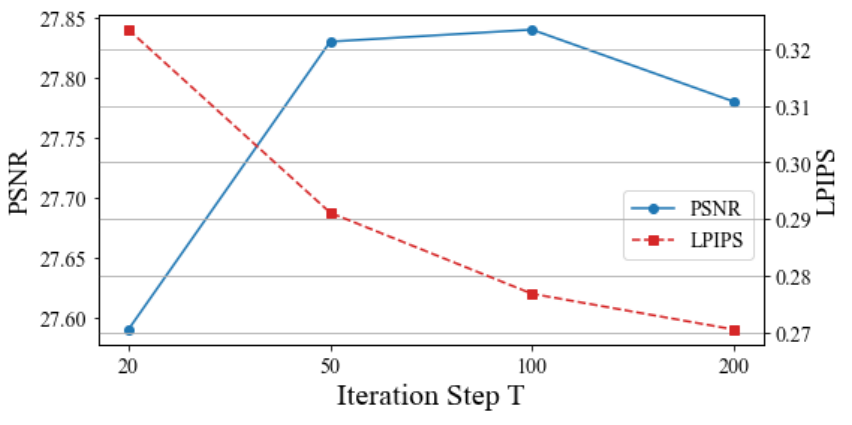}
\caption{Results of SR (4$\times$) on FFHQ dataset using different iteration step $T$.}
    \label{fig_iter_curve}
  \end{figure}

\section{Experiments}
\subsection{Experimental Setup}
\noindent\textbf{Datasets and tasks.} In this section, we examine the effectiveness of our proposed algorithm on two datasets: FFHQ 256$\times$256~\cite{karras2019style} and ImageNet 256$\times$256~\cite{deng2009imagenet}.
We conduct several IR tasks including \textit{Gaussian} deblur, \textit{motion} deblur, and super-resolution (SR).
Following~\cite{chung2023diffusion, zhu2023denoising}, the degradation is performed using 2$\times$, 4$\times$ and 8$\times$ bicubic down-sampling for SR.
For \textit{Gaussian} deblurring, a blur kernel of sized 61$\times$61 with a standard deviation of 3.0 is employed.
\siyuan{\textit{Motion deblurring} employs the same kernel size with an intensity of 0.5.}
%
Additive Gaussian noise with a variance of $\sigma_n=0.05$ is applied for
all degradation processes.

\noindent\textbf{Metrics.} In most IR tasks, both distortion metric and perceptual quality are significant for evaluating a reliable restoration.
Thus we employ Peak Signal-to-Noise Ratio (PSNR) and structural similarity (SSIM) as the distortion metrics to evaluate the fidelity of the restored results.
\siyuan{For perceptual quality, we use the Learned Perceptual Image Patch Similarity (LPIPS) distance~\cite{zhang2018unreasonable} for evaluation.}

\noindent\textbf{Implementation details.} The proposed method is implemented on PyTorch using NVIDIA RTX A5000. 
Following the previous work~\cite{zhu2023denoising}, the iteration step of our method is set to $T=100$ across each task and dataset.
\siyuan{More details on hyper-parameters are available in the appendix.}

\subsection{Comparison with State-of-the-Art}
To evaluate the performance of our proposed RDMD strategy, we conduct a comprehensive comparison with several state-of-the-art zero-shot IR methods.
Specifically, we adopt one PnP variants that perform deterministic regression, namely RED~\cite{romano2017little}.
For diffusion-based IR solvers, we \siyuan{select} MCG~\cite{chung2022improving}, DPS~\cite{chung2023diffusion}, DiffPIR~\cite{zhu2023denoising}, Resample~\cite{song2024solving}, and BIRD~\cite{chihaoui2024blind}.
For fair comparisons, we utilize the same pre-trained diffusion models for all methods, \siyuan{except for Resample, which leverages a pre-train latent diffusion model.}
For the FFHQ and ImageNet datasets, the pre-trained diffusion models are sourced from~\cite{chung2023diffusion} and~\cite{dhariwal2021diffusion}, respectively.
%
The pre-trained latent diffusion model for Resample is sourced from~\cite{rombach2022high}. 
The RED, DiffPIR, and our proposed method adopt the same iteration step $T=100$ for the sampling process.
In contrast, MCG and DPS necessitate $T=1000$ for DDPM sampling, and Resample requires $T=500$ steps.
Although BIRD employs only 10 steps for sampling, it necessitates fine-tuning over 200 epochs to optimize the initial noise map, which is equivalent to $T=2000$ steps.
%
%

%
As we propose a unified framework that blends the merit of both deterministic and stochastic strategies.
The existing methods RED~\cite{reehorst2018regularization} (Ours-deterministic) and DiffPIR~\cite{zhu2023denoising} (Ours-stochastic) can be regarded as the two special cases of our framework, as discussed in Section~\ref{sec_discussion}.

Tables~\ref{tab_ffhq} and ~\ref{tab_imagenet} present the quantitative results on the FFHQ and ImageNet datasets, respectively.
It can be found that the proposed method consistently outperforms or is highly competitive with other methods \siyuan{in various restoration tasks across both datasets.}
%
For the FFHQ dataset, which consists of homogeneous scenarios, our method could achieve superior results across almost all metrics, surpassing the existing state-of-the-art zero-shot IR methods by a significant margin.
By harnessing the merits of both deterministic and stochastic-based strategies, our method demonstrates a comparable or even better perceptual quality compared to pure stochastic methods (\eg DiffPIR), while presenting a significant boost in PSNR, especially in 8$\times$ SR with \siyuan{an increase of} +2.23dB.
%
\siyuan{For more diverse scenarios in the ImageNet dataset}
our approach significantly outperforms other existing zero-shot IR methods, especially in super-resolution tasks.
%

To further demonstrate the advantage of our method against other competing baselines, Figure~\ref{fig_visual} provides the qualitative results of super-resolution and deblurring on both the FFHQ and ImageNet datasets.
In this work, we consider a more practical yet challenging case involving addition Gaussian noise with $\sigma_n=0.05$, while MCG and Resample fail to deal with such noisy cases.
MCG leverages manifold constraints that project each intermediate result on the data manifold while leading to risky overfitting to the measurement noise.
A similar trend can also be observed in Resample, where the data consistency is enforced by minimizing the data term $\lVert \y - A\x\rVert^2_2$, leading to significant artifacts caused by measurement noise.
In contrast, by incorporating a deterministic regularizer into stochastic sampling, our method can effectively deal with such noise while maintaining a high perceptual quality, outperforming both deterministic and stochastic-based approaches.

\begin{figure} [t]
  \centering 
    \includegraphics[width=0.98\linewidth]{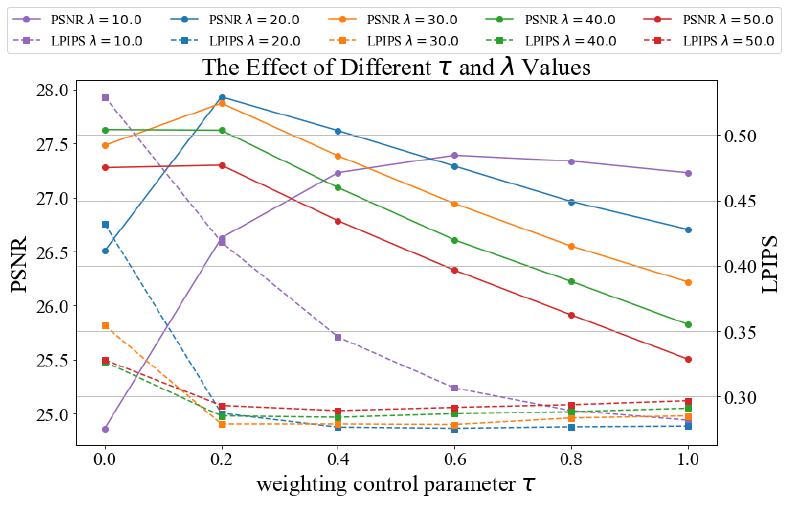}
\caption{Ablation study on the effect of different $\lambda$ and $\tau$ values on both PSNR and LPIPS in 4$\times$ super-resolution.}
  \vspace{-3mm}
    \label{fig_ablation_effect_lambda_tau}
  \end{figure}

\subsection{Ablation Study}
\noindent\textbf{The effect of iteration $T$.} 
We first explore the impact of the iteration count $T$ on performance.
We conduct experiments on 4$\times$ super-resolution for the FFHQ dataset using five distinct iteration steps: $[20, 50, 100, 200]$, as depicted in Figure~\ref{fig_iter_curve}.
%
The hyper-parameters are fixed across each experiment for a fair comparison.
As the iteration number $T$ increases, the PSNR values initially rise, reaching a peak around $T=100$ and then slightly decrease as $T$ continues to increase.
In contrast, the LPIPS scores generally improve with an increase in $T$.
According to Figure~\ref{fig_iter_curve}, our method demonstrates robust performance across a range of $T$ values, and we have chosen $T=100$ as the standard iteration count for all subsequent experiments.

\noindent\textbf{The effect of $\lambda$ and $\tau$.}
Our RDMD has two significant hyper-parameters $\lambda$ and $\tau$, where $\lambda$ controls the strength of the regularizer term compared to the data term, which is common in existing PnP works.
$\tau$ is a unique and key component in our work, which controls the relative contribution of deterministic and stochastic regularizers, thus enabling flexible customization of the distortion-perception tradeoff in the final restored results.
Figure~\ref{fig_ablation_effect_lambda_tau} illustrates the impact of varying the hyper-parameters $\lambda$ and $\tau$ on the performance of 4$\times$ super-resolution in the FFHQ dataset, with solid lines representing PSNR and dashed lines for LPIPS.
When $\lambda$ is too \siyuan{low} (\ie $\lambda=10$), our method fails to provide effective regularization on the restored results, leading to poor performance on both PSNR and LPIPS.
As $\lambda$ increases from 20 to 50, there is a noticeable decrease in PSNR across all $\tau$ values. 
This suggests that higher $\lambda$ values might lead to over-regularization, where the regularizer becomes too dominant, potentially smoothing out important details in the image and lowering the PSNR.
Unlike PSNR, LPIPS values tend to be better as $\tau$ increases, especially for $\lambda = 10$.
%
This trend suggests that incorporating stochastic regularization (\ie higher $\tau$) enhances perceptual quality by introducing diversity and variability that aligns with human perception.
We also observe consistent trends when the proposed method is applied to other tasks like deblurring, indicating that the trade-off between distortion and perception can be effectively controlled via these two parameters.
The capability of customizing this tradeoff is crucial for applications where either data fidelity or perceptual quality may be prioritized depending on the context.

\noindent\textbf{Comparison with dual network priors.}
In our framework, we only adopt a single diffusion model while using it in a dual-strategy.
Whereas, a more straightforward way is to directly leverage two networks (one discriminative and one generative) as the priors for regularization.
Thus, we conduct an ablation study to compare our method with two variants that take additional discriminative denoiser for regression.
Specifically, we pre-trained two popular backbones, namely DRUNet~\cite{zhang2021plug} and SwinIR~\cite{liang2021swinir}, to construct deterministic regularizers.
The results are provided in Table~\ref{tab_ablation_dualprior}, where \textit{Ours$+$DRUNet} denotes the variant of our framework using both pre-trained DRUNet and diffusion model, and similar for \textit{Ours$+$SwinIR}.
It demonstrates that our framework achieves comparable or even superior performance to dual-prior approaches in both PSNR and LPIPS, especially at lower and moderate upscaling factors (2$\times$ and 4$\times$ SR), without demanding extra networks.

\begin{table}
    \centering
    \setlength{\tabcolsep}{0.35mm}
    \caption{Ablation study on the comparison with dual network priors that use extra models to construct deterministic regularizer on FFHQ dataset.}
    \small
    \begin{tabular}{lcccccc}
    \toprule
         \textbf{FFHQ}&\multicolumn{2}{c}{\textbf{SR 2$\times$}}& \multicolumn{2}{c}{\textbf{SR 4$\times$}}& \multicolumn{2}{c}{\textbf{SR 8$\times$}}\\
     \midrule
        Model &PSNR$\uparrow$&LPIPS$\downarrow$&PSNR$\uparrow$&LPIPS$\downarrow$&PSNR$\uparrow$&LPIPS$\downarrow$\\
        \midrule
         Ours$+$DRUNet& 30.77 & 0.2049 & 27.66 & 0.2890 & \textbf{23.34} & \textbf{0.3603}\\
         Ours$+$SwinIR& 30.85 & 0.2021 & 27.69 & 0.2903 & 23.33 & 0.3618 \\
         \midrule
         \cellcolor{gray!20}Ours&\cellcolor{gray!20}\textbf{31.03} &\cellcolor{gray!20}\textbf{0.1807} &\cellcolor{gray!20}\textbf{27.84} &\cellcolor{gray!20}\textbf{0.2768} &\cellcolor{gray!20}23.31 &\cellcolor{gray!20}0.3614\\ 
      \bottomrule
    \end{tabular}
    \label{tab_ablation_dualprior}
    \vspace{-2mm}
\end{table}

\section{Conlcusion}
As the data fidelity or perceptual quality may be prioritized varying from different image restoration tasks.
Possessing the capability to achieve a distortion-perception tradeoff is crucial for effectively addressing various IR problems.
In this work, we show that a single diffusion model can be used in a dual-strategy (\ie stochastic and deterministic) to construct two complementary regularizers.
Building upon this, we conduct the first attempt to integrate
the deterministic and stochastic strategies in a unified framework that solves zero-shot IR problems iteratively.
Such a unified framework enables flexible customization of the
the distortion-perception tradeoff, enhancing its suitability for a
variety of distinct IR tasks. 
Finally, extensive results on several IR tasks demonstrate the superior performance of the proposed methods on both FFHQ and ImageNet datasets.

{
    \small
    \bibliographystyle{ieeenat_fullname}
    \bibliography{main}

\begin{thebibliography}{41}
\providecommand{\natexlab}[1]{#1}
\providecommand{\url}[1]{\texttt{#1}}
\expandafter\ifx\csname urlstyle\endcsname\relax
  \providecommand{\doi}[1]{doi: #1}\else
  \providecommand{\doi}{doi: \begingroup \urlstyle{rm}\Url}\fi

\bibitem[Afonso et~al.(2010)Afonso, Bioucas-Dias, and Figueiredo]{hqs}
M.~V Afonso, J.~M Bioucas-Dias, and M.~AT Figueiredo.
\newblock Fast image recovery using variable splitting and constrained optimization.
\newblock \emph{IEEE transactions on image processing}, 19\penalty0 (9):\penalty0 2345--2356, 2010.

\bibitem[Aharon et~al.(2006)Aharon, Elad, and Bruckstein]{aharon2006k}
Michal Aharon, Michael Elad, and Alfred Bruckstein.
\newblock K-svd: An algorithm for designing overcomplete dictionaries for sparse representation.
\newblock \emph{IEEE Transactions on signal processing}, 54\penalty0 (11):\penalty0 4311--4322, 2006.

\bibitem[Blau and Michaeli(2018)]{blau2018perception}
Yochai Blau and Tomer Michaeli.
\newblock The perception-distortion tradeoff.
\newblock In \emph{Proceedings of the IEEE conference on computer vision and pattern recognition}, pages 6228--6237, 2018.

\bibitem[Boyd et~al.(2011)Boyd, Parikh, Chu, Peleato, Eckstein, et~al.]{boyd2011distributed}
Stephen Boyd, Neal Parikh, Eric Chu, Borja Peleato, Jonathan Eckstein, et~al.
\newblock Distributed optimization and statistical learning via the alternating direction method of multipliers.
\newblock \emph{Foundations and Trends{\textregistered} in Machine learning}, 3\penalty0 (1):\penalty0 1--122, 2011.

\bibitem[Chan et~al.(2016)Chan, Wang, and Elgendy]{chan2016plug}
Stanley~H Chan, Xiran Wang, and Omar~A Elgendy.
\newblock Plug-and-play admm for image restoration: Fixed-point convergence and applications.
\newblock \emph{IEEE Transactions on Computational Imaging}, 3\penalty0 (1):\penalty0 84--98, 2016.

\bibitem[Chihaoui et~al.(2024)Chihaoui, Lemkhenter, and Favaro]{chihaoui2024blind}
Hamadi Chihaoui, Abdelhak Lemkhenter, and Paolo Favaro.
\newblock Blind image restoration via fast diffusion inversion.
\newblock \emph{arXiv preprint arXiv:2405.19572}, 2024.

\bibitem[Chung et~al.(2022)Chung, Sim, Ryu, and Ye]{chung2022improving}
Hyungjin Chung, Byeongsu Sim, Dohoon Ryu, and Jong~Chul Ye.
\newblock Improving diffusion models for inverse problems using manifold constraints.
\newblock \emph{Advances in Neural Information Processing Systems}, 35:\penalty0 25683--25696, 2022.

\bibitem[Chung et~al.(2023)Chung, Kim, Mccann, Klasky, and Ye]{chung2023diffusion}
Hyungjin Chung, Jeongsol Kim, Michael~T Mccann, Marc~L Klasky, and Jong~Chul Ye.
\newblock Diffusion posterior sampling for general noisy inverse problems.
\newblock In \emph{The Eleventh International Conference on Learning Representations, ICLR 2023}. The International Conference on Learning Representations, 2023.

\bibitem[Dabov et~al.(2007)Dabov, Foi, Katkovnik, and Egiazarian]{dabov2007image}
Kostadin Dabov, Alessandro Foi, Vladimir Katkovnik, and Karen Egiazarian.
\newblock Image denoising by sparse 3-d transform-domain collaborative filtering.
\newblock \emph{IEEE Transactions on image processing}, 16\penalty0 (8):\penalty0 2080--2095, 2007.

\bibitem[Deng et~al.(2009)Deng, Dong, Socher, Li, Li, and Fei-Fei]{deng2009imagenet}
Jia Deng, Wei Dong, Richard Socher, Li-Jia Li, Kai Li, and Li Fei-Fei.
\newblock Imagenet: A large-scale hierarchical image database.
\newblock In \emph{2009 IEEE conference on computer vision and pattern recognition}, pages 248--255. Ieee, 2009.

\bibitem[Dhariwal and Nichol(2021)]{dhariwal2021diffusion}
Prafulla Dhariwal and Alexander Nichol.
\newblock Diffusion models beat gans on image synthesis.
\newblock \emph{Advances in Neural Information Processing Systems}, 34:\penalty0 8780--8794, 2021.

\bibitem[Efron(2011)]{efron2011tweedie}
Bradley Efron.
\newblock Tweedie’s formula and selection bias.
\newblock \emph{Journal of the American Statistical Association}, 106\penalty0 (496):\penalty0 1602--1614, 2011.

\bibitem[Freirich et~al.(2021)Freirich, Michaeli, and Meir]{freirich2021theory}
Dror Freirich, Tomer Michaeli, and Ron Meir.
\newblock A theory of the distortion-perception tradeoff in wasserstein space.
\newblock \emph{Advances in Neural Information Processing Systems}, 34:\penalty0 25661--25672, 2021.

\bibitem[Ho et~al.(2020)Ho, Jain, and Abbeel]{ho2020denoising}
Jonathan Ho, Ajay Jain, and Pieter Abbeel.
\newblock Denoising diffusion probabilistic models.
\newblock \emph{Advances in neural information processing systems}, 33:\penalty0 6840--6851, 2020.

\bibitem[Ho et~al.(2022)Ho, Saharia, Chan, Fleet, Norouzi, and Salimans]{ho2022cascaded}
Jonathan Ho, Chitwan Saharia, William Chan, David~J Fleet, Mohammad Norouzi, and Tim Salimans.
\newblock Cascaded diffusion models for high fidelity image generation.
\newblock \emph{J. Mach. Learn. Res.}, 23\penalty0 (47):\penalty0 1--33, 2022.

\bibitem[Karras et~al.(2019)Karras, Laine, and Aila]{karras2019style}
Tero Karras, Samuli Laine, and Timo Aila.
\newblock A style-based generator architecture for generative adversarial networks.
\newblock In \emph{Proceedings of the IEEE/CVF conference on computer vision and pattern recognition}, pages 4401--4410, 2019.

\bibitem[Liang et~al.(2021)Liang, Cao, Sun, Zhang, Van~Gool, and Timofte]{liang2021swinir}
Jingyun Liang, Jiezhang Cao, Guolei Sun, Kai Zhang, Luc Van~Gool, and Radu Timofte.
\newblock Swinir: Image restoration using swin transformer.
\newblock In \emph{Proceedings of the IEEE/CVF international conference on computer vision}, pages 1833--1844, 2021.

\bibitem[Lin et~al.(2023)Lin, He, Chen, Lyu, Fei, Dai, Ouyang, Qiao, and Dong]{lin2023diffbir}
Xinqi Lin, Jingwen He, Ziyan Chen, Zhaoyang Lyu, Ben Fei, Bo Dai, Wanli Ouyang, Yu Qiao, and Chao Dong.
\newblock Diffbir: Towards blind image restoration with generative diffusion prior.
\newblock \emph{arXiv preprint arXiv:2308.15070}, 2023.

\bibitem[Liu et~al.(2019)Liu, Zhang, and Xiong]{liu2019classification}
Dong Liu, Haochen Zhang, and Zhiwei Xiong.
\newblock On the classification-distortion-perception tradeoff.
\newblock \emph{Advances in Neural Information Processing Systems}, 32, 2019.

\bibitem[Nichol and Dhariwal(2021)]{nichol2021improved}
Alexander~Quinn Nichol and Prafulla Dhariwal.
\newblock Improved denoising diffusion probabilistic models.
\newblock In \emph{International Conference on Machine Learning}, pages 8162--8171. PMLR, 2021.

\bibitem[Reehorst and Schniter(2018)]{reehorst2018regularization}
Edward~T Reehorst and Philip Schniter.
\newblock Regularization by denoising: Clarifications and new interpretations.
\newblock \emph{IEEE transactions on computational imaging}, 5\penalty0 (1):\penalty0 52--67, 2018.

\bibitem[Robbins(1992)]{robbins1992empirical}
Herbert~E Robbins.
\newblock An empirical bayes approach to statistics.
\newblock In \emph{Breakthroughs in Statistics: Foundations and basic theory}, pages 388--394. Springer, 1992.

\bibitem[Romano et~al.(2017)Romano, Elad, and Milanfar]{romano2017little}
Yaniv Romano, Michael Elad, and Peyman Milanfar.
\newblock The little engine that could: Regularization by denoising (red).
\newblock \emph{SIAM Journal on Imaging Sciences}, 10\penalty0 (4):\penalty0 1804--1844, 2017.

\bibitem[Rombach et~al.(2022)Rombach, Blattmann, Lorenz, Esser, and Ommer]{rombach2022high}
Robin Rombach, Andreas Blattmann, Dominik Lorenz, Patrick Esser, and Bj{\"o}rn Ommer.
\newblock High-resolution image synthesis with latent diffusion models.
\newblock In \emph{Proceedings of the IEEE/CVF Conference on Computer Vision and Pattern Recognition}, pages 10684--10695, 2022.

\bibitem[Shang et~al.(2024)Shang, Shan, Liu, Wang, Wang, Zhang, and Zhang]{shang2024resdiff}
Shuyao Shang, Zhengyang Shan, Guangxing Liu, LunQian Wang, XingHua Wang, Zekai Zhang, and Jinglin Zhang.
\newblock Resdiff: Combining cnn and diffusion model for image super-resolution.
\newblock In \emph{Proceedings of the AAAI Conference on Artificial Intelligence}, number~8, pages 8975--8983, 2024.

\bibitem[Sohl-Dickstein et~al.(2015)Sohl-Dickstein, Weiss, Maheswaranathan, and Ganguli]{sohl2015deep}
Jascha Sohl-Dickstein, Eric Weiss, Niru Maheswaranathan, and Surya Ganguli.
\newblock Deep unsupervised learning using nonequilibrium thermodynamics.
\newblock In \emph{International Conference on Machine Learning}, pages 2256--2265. PMLR, 2015.

\bibitem[Song et~al.(2024)Song, Kwon, Zhang, Hu, Qu, and Shen]{song2024solving}
Bowen Song, Soo~Min Kwon, Zecheng Zhang, Xinyu Hu, Qing Qu, and Liyue Shen.
\newblock Solving inverse problems with latent diffusion models via hard data consistency.
\newblock In \emph{The Twelfth International Conference on Learning Representations}, 2024.

\bibitem[Song et~al.(2020)Song, Meng, and Ermon]{song2020denoising}
Jiaming Song, Chenlin Meng, and Stefano Ermon.
\newblock Denoising diffusion implicit models.
\newblock \emph{arXiv preprint arXiv:2010.02502}, 2020.

\bibitem[Stein(1981)]{stein1981estimation}
Charles~M Stein.
\newblock Estimation of the mean of a multivariate normal distribution.
\newblock \emph{The annals of Statistics}, pages 1135--1151, 1981.

\bibitem[Venkatakrishnan et~al.(2013)Venkatakrishnan, Bouman, and Wohlberg]{venkatakrishnan2013plug}
Singanallur~V Venkatakrishnan, Charles~A Bouman, and Brendt Wohlberg.
\newblock Plug-and-play priors for model based reconstruction.
\newblock In \emph{2013 IEEE global conference on signal and information processing}, pages 945--948. IEEE, 2013.

\bibitem[Wang et~al.(2022)Wang, Zhang, Ravishankar, and Wen]{wang2022repnp}
Chong Wang, Rongkai Zhang, Saiprasad Ravishankar, and Bihan Wen.
\newblock Repnp: Plug-and-play with deep reinforcement learning prior for robust image restoration.
\newblock In \emph{2022 IEEE International Conference on Image Processing (ICIP)}, pages 2886--2890. IEEE, 2022.

\bibitem[Wang et~al.(2023{\natexlab{a}})Wang, Zhang, Maliakal, Ravishankar, and Wen]{wang2023deep}
Chong Wang, Rongkai Zhang, Gabriel Maliakal, Saiprasad Ravishankar, and Bihan Wen.
\newblock Deep reinforcement learning based unrolling network for mri reconstruction.
\newblock In \emph{2023 IEEE 20th International Symposium on Biomedical Imaging (ISBI)}, pages 1--5. IEEE, 2023{\natexlab{a}}.

\bibitem[Wang et~al.(2024)Wang, Guo, Wang, Cheng, Yu, and Wen]{wang2024progressive}
Chong Wang, Lanqing Guo, Yufei Wang, Hao Cheng, Yi Yu, and Bihan Wen.
\newblock Progressive divide-and-conquer via subsampling decomposition for accelerated mri.
\newblock In \emph{Proceedings of the IEEE/CVF Conference on Computer Vision and Pattern Recognition}, pages 25128--25137, 2024.

\bibitem[Wang et~al.(2023{\natexlab{b}})Wang, Yu, and Zhang]{wang2022zero}
Yinhuai Wang, Jiwen Yu, and Jian Zhang.
\newblock Zero-shot image restoration using denoising diffusion null-space model.
\newblock \emph{The Eleventh International Conference on Learning Representations}, 2023{\natexlab{b}}.

\bibitem[Wang et~al.(2023{\natexlab{c}})Wang, Zhang, Zhang, Zheng, Zhou, Zhang, and Wang]{wang2023dr2}
Zhixin Wang, Ziying Zhang, Xiaoyun Zhang, Huangjie Zheng, Mingyuan Zhou, Ya Zhang, and Yanfeng Wang.
\newblock Dr2: Diffusion-based robust degradation remover for blind face restoration.
\newblock In \emph{Proceedings of the IEEE/CVF Conference on Computer Vision and Pattern Recognition}, pages 1704--1713, 2023{\natexlab{c}}.

\bibitem[Wen et~al.(2020)Wen, Li, Li, and Bresler]{wen2020set}
Bihan Wen, Yanjun Li, Yuqi Li, and Yoram Bresler.
\newblock A set-theoretic study of the relationships of image models and priors for restoration problems.
\newblock \emph{arXiv preprint arXiv:2003.12985}, 2020.

\bibitem[Zha et~al.(2023)Zha, Wen, Yuan, Zhang, Zhou, Jiang, and Zhu]{zha2023multiple}
Zhiyuan Zha, Bihan Wen, Xin Yuan, Jiachao Zhang, Jiantao Zhou, Xudong Jiang, and Ce Zhu.
\newblock Multiple complementary priors for multispectral image compressive sensing reconstruction.
\newblock \emph{IEEE Transactions on Cybernetics}, 2023.

\bibitem[Zhang et~al.(2017)Zhang, Zuo, Gu, and Zhang]{zhang2017learning}
Kai Zhang, Wangmeng Zuo, Shuhang Gu, and Lei Zhang.
\newblock Learning deep cnn denoiser prior for image restoration.
\newblock In \emph{Proceedings of the IEEE conference on computer vision and pattern recognition}, pages 3929--3938, 2017.

\bibitem[Zhang et~al.(2021)Zhang, Li, Zuo, Zhang, Van~Gool, and Timofte]{zhang2021plug}
Kai Zhang, Yawei Li, Wangmeng Zuo, Lei Zhang, Luc Van~Gool, and Radu Timofte.
\newblock Plug-and-play image restoration with deep denoiser prior.
\newblock \emph{IEEE Transactions on Pattern Analysis and Machine Intelligence}, 44\penalty0 (10):\penalty0 6360--6376, 2021.

\bibitem[Zhang et~al.(2018)Zhang, Isola, Efros, Shechtman, and Wang]{zhang2018unreasonable}
Richard Zhang, Phillip Isola, Alexei~A Efros, Eli Shechtman, and Oliver Wang.
\newblock The unreasonable effectiveness of deep features as a perceptual metric.
\newblock In \emph{Proceedings of the IEEE conference on computer vision and pattern recognition}, pages 586--595, 2018.

\bibitem[Zhu et~al.(2023)Zhu, Zhang, Liang, Cao, Wen, Timofte, and Van~Gool]{zhu2023denoising}
Yuanzhi Zhu, Kai Zhang, Jingyun Liang, Jiezhang Cao, Bihan Wen, Radu Timofte, and Luc Van~Gool.
\newblock Denoising diffusion models for plug-and-play image restoration.
\newblock In \emph{Proceedings of the IEEE/CVF Conference on Computer Vision and Pattern Recognition}, pages 1219--1229, 2023.

\end{thebibliography}
}


\end{document}